%% file: main.tex
\documentclass[letterpaper, 10 pt, conference]{ieeeconf}  

\IEEEoverridecommandlockouts                              

\usepackage[T1]{fontenc}
\title{\LARGE \bf
Incorporating Heterophily into Graph Neural Networks for Graph Classification
}
\usepackage{color,array}

\usepackage{soul}
\usepackage{url}
\usepackage{caption}
\usepackage{graphicx}
\usepackage{amsmath}

\usepackage{amsfonts}
\usepackage{amssymb}
\usepackage{booktabs}
\usepackage{algorithm}
\usepackage{algorithmic}
\usepackage[switch]{lineno}
\usepackage{color}
\usepackage[table]{xcolor}
\usepackage{dsfont}

\usepackage{chemfig}

\usepackage{mathtools}

\usepackage{multirow}
\usepackage{float}
\usepackage[ruled, vlined,algo2e,linesnumbered]{algorithm2e}
\usepackage{tikz}
\usepackage{xcolor}

\usepackage{tkz-euclide,subfigure}
\usetikzlibrary{shapes}
\newlength{\R}
\usepackage{pgfplots}
\pgfplotsset{compat=newest}
\usetikzlibrary{shapes.arrows}
\usetikzlibrary{arrows.meta}
\usetikzlibrary{trees,arrows}

\newcommand{\nosemic}{\renewcommand{\@endalgocfline}{\relax}}
\newcommand{\dosemic}{\renewcommand{\@endalgocfline}{\algocf@endline}}

\newcommand\footnoteref[1]{\protected@xdef\@thefnmark{\ref{#1}}\@footnotemark}

\newcommand{\RN}[1]{%
	\textup{\uppercase\expandafter{\romannumeral#1}}%
}

\newcommand{\readout}{\mathop{\mathrm{READOUT}}}

\newcommand*{\ldblbrace}{\left\lbrace \mskip-7mu\left\lbrace }
\newcommand*{\rdblbrace}{\right\rbrace \mskip-7mu\right\rbrace }

\author{Jiayi Yang$^{1}$, Sourav Medya$^{2}$, and Wei Ye$^{1,\ast}$
\thanks{$^{1}$Jiayi Yang and Wei Ye are with the College of Electronic and Information Engineering, Tongji University, Shanghai 201804 China
        {\tt\small \{2111125, yew\}@tongji.edu.cn}}%
\thanks{$^{2}$Sourav Medya is with the Department of Computer Science, University of Illinois, Chicago, IL 60607 USA
        {\tt\small medya@uic.edu}}%
}

\begin{document}

\maketitle
\thispagestyle{empty}
\pagestyle{empty}

\begingroup\renewcommand\thefootnote{$^\ast$}
\footnotetext{Corresponding author.}
\endgroup

\begin{abstract}

Graph Neural Networks (GNNs) often assume strong homophily for graph classification, seldom considering heterophily, which means connected nodes tend to have different class labels and dissimilar features. In real-world scenarios, graphs may have nodes that exhibit both homophily and heterophily. Failing to generalize to this setting makes many GNNs underperform in graph classification. In this paper, we address this limitation by identifying three effective designs and develop a novel GNN architecture called IHGNN (\textbf{\underline{I}}ncorporating \textbf{\underline{H}}eterophily into \textbf{\underline{G}}raph \textbf{\underline{N}}eural \textbf{\underline{N}}etworks). These designs include the combination of integration and separation of the ego- and neighbor-embeddings of nodes, adaptive aggregation of node embeddings from different layers, and differentiation between different node embeddings for constructing the graph-level readout function. We empirically validate IHGNN on various graph datasets and demonstrate that it outperforms the state-of-the-art GNNs for graph classification.

\end{abstract}

\input{sec_introduction}

\input{sec_relatedwork}
\input{sec_preliminaries}

\input{sec_methods}

\input{sec_experiments}

\input{sec_conclusion}






\bibliographystyle{plain}
\bibliography{reference}

\end{document}

%% file: sec_introduction.tex
\section{INTRODUCTION}\label{sec:intro}

Graphs can encode and represent relational structures that appear in many domains and thus are widely used in real-world applications. 
In the past years, graph neural networks (GNNs)~\cite{scarselli2008graph,gilmer2017neural,defferrard2016convolutional,kipf2016semi} have received a lot of attention as they are effective in learning representations of nodes as well as graphs and solving supervised learning problems on graphs.

In the real world, graphs may exhibit both homophily and heterophily. Basically, homophily means the connected nodes tend to have the same class label and similar features (``birds of a feather flock together'')~\cite{mcpherson2001birds} while 
heterophily means the nodes having different class labels and dissimilar features are more likely to link together.
Most GNNs follow the strong-homophily-assumed iterative message passing~\cite{gilmer2017neural} scheme (MPNNs), where node embeddings (i.e., messages) are uniformly aggregated and propagated through the graph structure, e.g., ``average'' in vanilla GCN~\cite{kipf2016semi} and ``sum'' in GIN~\cite{xu2018powerful}. However, recent works~\cite{zhu2020beyond,zheng2022graph,suresh2021breaking,yan2022two,wu2024dcgnn} have shown that using such uniform aggregation in graphs in which heterophily dominates homophily may lead to performance degradation. The essential intuition behind this lies in that the fusion of heterophilic information from neighbors could diminish the distinguishability of original features and may even misclassify the node under certain circumstances. 

To handle heterophily in graphs, many researchers try to modify the aggregation and updating steps in GNNs~\cite{pei2020geom,zhu2020beyond} or explain the message-passing mechanism 
from the perspective of graph signal processing~\cite{luan2022revisiting,li2024pc,bo2021beyond}. H$_2$GCN~\cite{zhu2020beyond} uses skip connection to aggregate the ego- and neighbor-embedding separately. 
FAGCN~\cite{bo2021beyond} and ACM~\cite{luan2022revisiting} additionally use high-frequency information in the aggregation, aiming to capture the difference between nodes, in other words, heterophily. 

\begin{figure}[t]
\centering
\subfigure[MUTAG (Ave. Hom. ratio 0.65)]
{
    \begin{minipage}[b]{.46\linewidth}
        \centering
        \includegraphics[scale=0.20]{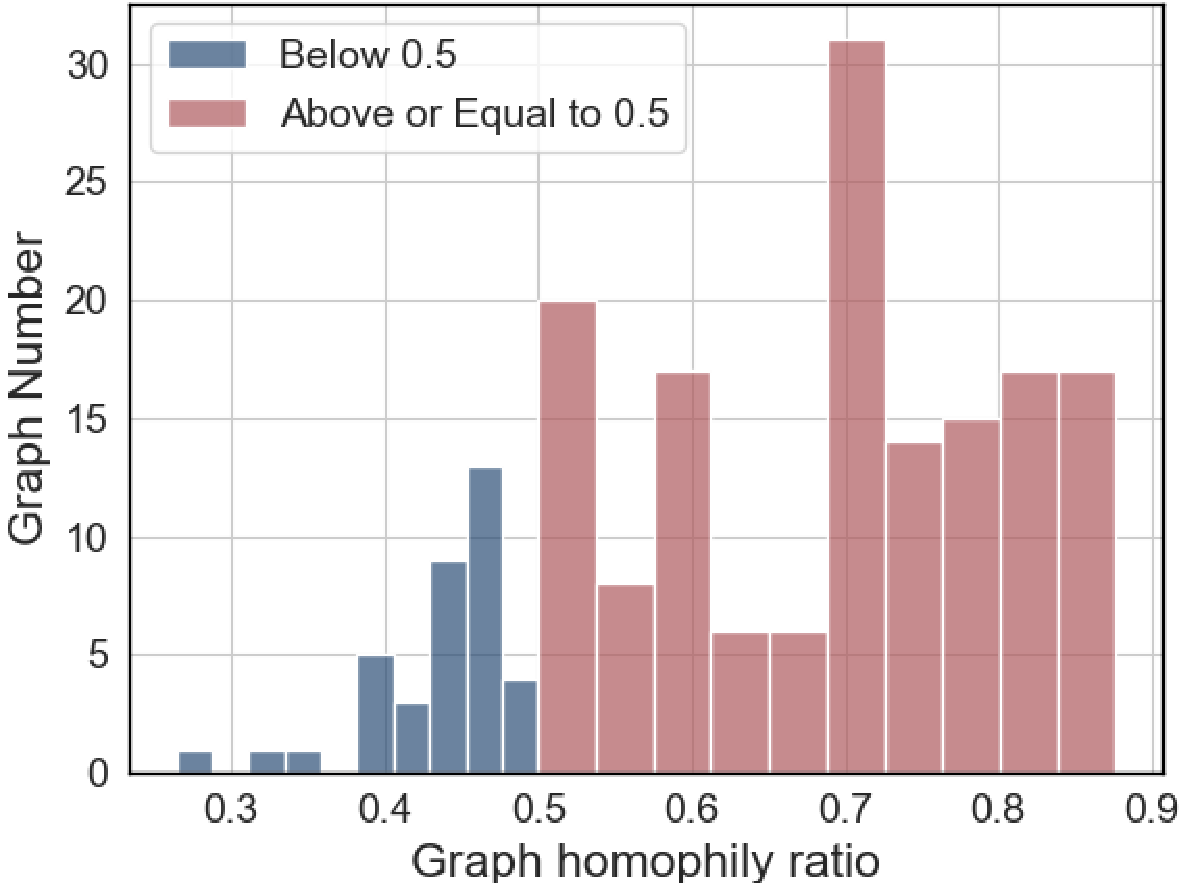}
    \end{minipage}
}
\subfigure[COLLAB (Ave. Hom. ratio 0.30)]
{
 	\begin{minipage}[b]{.46\linewidth}
        \centering
        \includegraphics[scale=0.20]{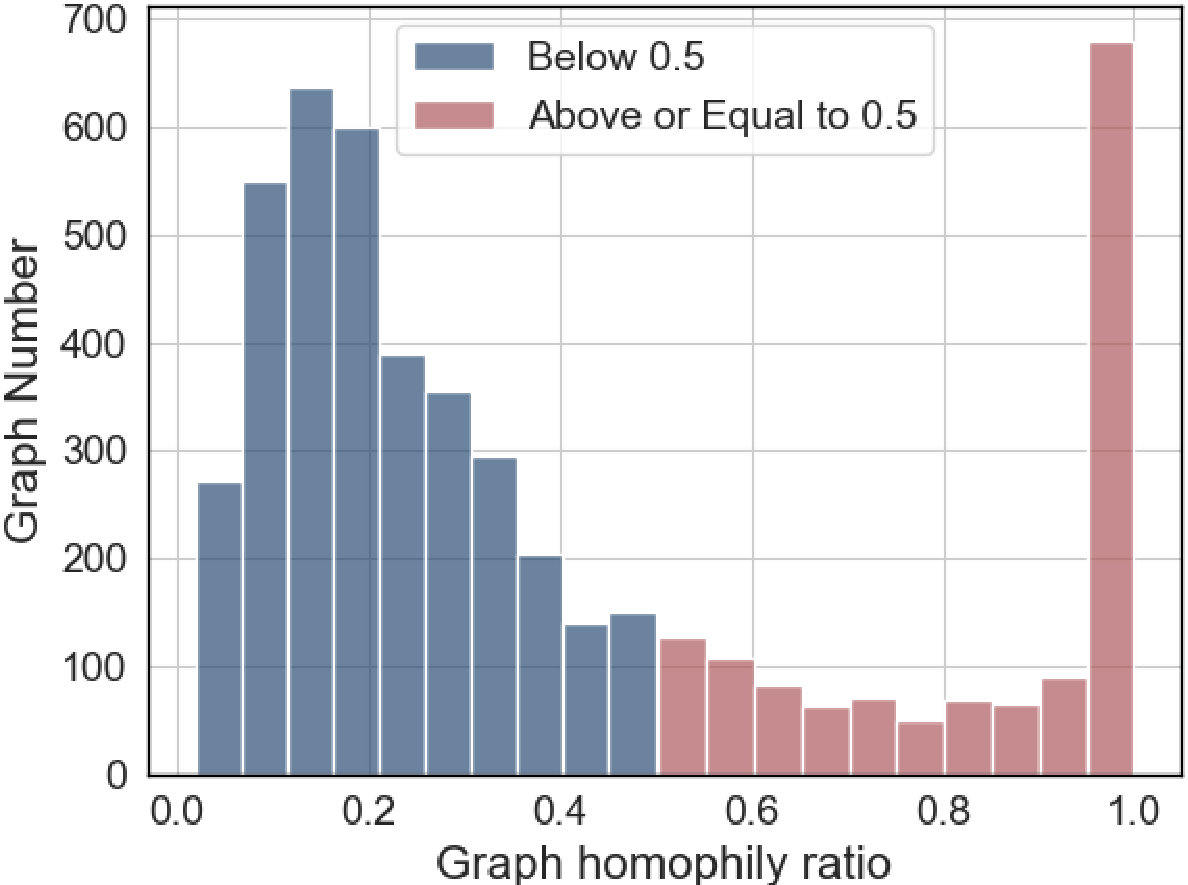}
    \end{minipage}
}

\caption{
The histogram of the graph homophily ratios for the MUTAG and the COLLAB dataset.
} 
\label{homo_ratio}
\end{figure}

However, all the works above focus on the node classification task. The graph classification task, on the other hand, is another major task in graph applications and aims to classify graphs into different categories. For graph classification, GNNs involve an additional readout module~\cite{zhang2018end,feng2022kergnns,pasa2022polynomial,wu2024dcgnn} or use a hierarchically pooling operation~\cite{ying2018hierarchical,yuan2020structpool,yang2021hierarchical,liu2022unsupervised} to generate graph representations. Although summation has been proven to be a straightforward and powerful approach as the readout function~\cite{xu2018powerful}, it might not be adequate for graph classification when graphs have both homophily and heterophily. As shown in Fig.~\ref{homo_ratio}, the COLLAB dataset has an average homophily ratio of 0.3, which indicates that COLLAB contains more heterophily. Nevertheless, it still contains graphs whose homophily is close to 1. The MUTAG dataset containing more homophily exhibits similar characteristics.
In contrast to the node classification task, which only needs to emphasize a single graph with a certain homophily ratio, the graph classification task requires adaptation to numerous graphs with varying homophily ratios, spanning a wide range. Using uniform aggregation alone is not adaptive to graphs with different homophily ratios and may yield low performance. In this regard, the readout function should take the difference in homophily ratio across graphs into consideration.

In this paper, we develop a novel graph neural network model called IHGNN (\textbf{\underline{I}}ncorporating \textbf{\underline{H}}eterophily into \textbf{\underline{G}}raph \textbf{\underline{N}}eural \textbf{\underline{N}}etworks) for graph classification. Firstly, we propose an intra-layer message-passing mechanism to concurrently integrate and separate the ego- and neighbor-embedding. This allows our method to be adaptive to graphs with varying homophily ratios at the node-level. 
Secondly, we adaptively aggregate the inter-layer information to leverage high-order neighbors, recognizing that similar nodes are typically situated at long distances in graphs where heterophily dominates homophily. Finally, IHGNN designs a permutation-invariant and size-invariant readout function to reorganize the nodes within each graph and align them across the graph, thereby enabling our method to address graphs with varying homophily ratios at the graph-level.
Our main contributions can be summarized as follows:
\begin{itemize}
	\item We propose to use both the integration and separation of the ego- and neighbor-embeddings of nodes in the $\mbox{COMBINE}(\cdot)$ operator (formalized in equation~(\ref{eqn:agg_com}) in Section~\ref{sec:gnn}) of GNNs. 
 The new $\mbox{COMBINE}(\cdot)$ operator makes GNNs adaptive to different graphs with varying homophily ratios.
	\item We propose to adaptively aggregate the intermediate node embeddings from different layers by Single-Layer Perceptrons (SLP) and use the Multi-layer Perceptrons (MLP) to model the $\mbox{COMBINE}(\cdot)$ operator in GNNs.
 
	\item 
	We propose a graph-level readout function that discerns each node embedding, i.e., the embeddings of nodes with different homophily ratios are not mixed together.
 This is achieved by aligning nodes from different graphs in the same order as performed by the continuous 1-WL~\cite{leman1968reduction} algorithm for $K$ iterations.
	\item We develop a new GNN model called IHGNN that adopts all the above designs. Empirical results show that IHGNN outperforms other state-of-the-art GNNs on various graph datasets in terms of classification accuracy.
\end{itemize}

%% file: sec_relatedwork.tex
\section{RELATED WORK}\label{relatedWork}
\subsection{GNNs for Graph Classification}

Many GNN models such as~\cite{defferrard2016convolutional,kipf2016semi,xu2018representation,chamberlain2021grand,velivckovic2017graph,hamilton2017inductive} follow a common framework called message passing neural networks (MPNNs). This framework entails the aggregation and updating of node representations, followed by a readout function to obtain the graph representations.
In contrast to node classification, which is a node-level task and thus only requires the aggregation and updating steps of MPNNs, graph classification employs different permutation-invariant and size-invariant readout functions to generate graph-level representations~\cite{zhang2018end,feng2022kergnns,pasa2022polynomial,wu2024dcgnn} for the graph-structured inputs whose size and topology are varying. 

DGCNN~\cite{zhang2018end} proposes to use the random walk probability transition matrix as graph Laplacian and relates itself to the Weisfeiler-Lehman subtree kernel~\cite{shervashidze2009fast,shervashidze2011weisfeiler} and propagation kernel~\cite{neumann2016propagation}.
KerGNNs~\cite{feng2022kergnns} integrates trainable graph kernels (random walk graph kernel~\cite{gartner2003graph,kashima2003marginalized}) into a subgraph-based message passing process and uses sum pooling as the readout function.
PGCN~\cite{pasa2022polynomial} proposes a new convolution operator with a larger receptive field, which consists of a polynomial series of powers of the graph adjacency matrix. The readout component of PGCN consists of a concatenation of three different aggregation strategies: sum, mean, and max.

Other methods~\cite{ying2018hierarchical,yuan2020structpool,yang2021hierarchical,liu2022unsupervised} 
coarsen the input graph hierarchically into smaller graphs using graph pooling.
DiffPool~\cite{ying2018hierarchical} is a differentiable graph pooling module that learns a soft assignment of each node to a set of clusters at each layer. Each cluster corresponds to the coarsened input for the next layer. 
At each layer, DiffPool uses two separate GNNs to generate the embeddings and assignment matrices without considering the relations between them. 

Several works have been devoted to relating GNNs to the Weisfeiler-Lehman graph isomorphism test~\cite{leman1968reduction}. GIN~\cite{xu2018powerful} proposes to use MLP to learn the injective multiset functions for the neighbor aggregation. $k$-GNNs~\cite{morris2019weisfeiler} considers higher-order graph structures at multiple scales and performs message passing directly between graph substructures rather than individual nodes. PPGNs~\cite{maron2019provably} apply MLP to the feature dimension and matrix multiplication for developing scalable GNNs. GNNML~\cite{balcilar2021breaking} designs a new GNN architecture that is based on the recently proposed Matrix Language called MATLANG~\cite{brijder2019expressive,geerts2021expressive}. However, all these models perform message passing between higher-order relations in graphs, which leads to computational overhead. Besides, they do not consider both the homophily and heterophily in graphs, and thus their architectures are not adaptive to graphs with varying node homophily ratios.

Recently, researchers have developed some Transformers~\cite{ying2021transformers,kim2022pure,diao2022relational} for graph representation learning. However, they are more suitable for large-scale graph tasks, such as shortest paths task~\cite{tang2020towards} and OGB-LSC~\cite{hu2103ogb} quantum chemistry regression.


\subsection{GNNs Addressing Heterophily}

Most of the above GNNs deal with homophilous graphs, whose connected nodes tend to have the same class label. However, in heterophilous graphs, connected nodes tend to have different class labels and dissimilar features. Homophilous GNNs do not perform well in this scenario. 
To mitigate the limitations, Geom-GCN~\cite{pei2020geom} proposes a new geometric aggregation scheme that includes three modules, i.e., node embedding, structural neighborhood, and bi-level aggregation. H$_2$GCN~\cite{zhu2020beyond} identifies three designs to improve learning on heterophilous graphs. 
GloGNN~\cite{li2022finding} treats the aggregation as a new optimization problem in a propagation-decoupled way and finds a closed-form solution for it. 

Looking through the lens of graph signal processing, some methods use graph filters to learn high-pass and low-pass graph signals, which indicate heterophily and homophily, respectively. GPR-GNN~\cite{chien2020adaptive} proposes a new generalized pagerank (GPR) GNN architecture that corresponds to a polynomial graph filtering of order $K$ and learns the heterophily by allowing the GPR weights to be negative. ASGC~\cite{chanpuriya2022simplified} calculates the GPR weights based on linear least squares. ACM~\cite{luan2022revisiting} combines four different forms of Laplacian matrices as high-pass filters and affinity matrices as low-pass filters which can prevent the loss of input information and learn different weights for each node in different filters. 
PCNet~\cite{li2024pc} proposes a two-fold filtering mechanism that particularly uses a graph heat equation as one filter to aggregate the heterophilous global information.
All the above GNNs are developed for node-level tasks such as node classification. By summing all the node embeddings, they can be extended to graph-level tasks such as graph classification. 

The work tailored for classifying graphs with heterophily is much less explored.
DCGNN~\cite{wu2024dcgnn} focuses on scenarios that struggle to discern circular substructures that are prevalent in the heterophily and utilizes a tree-shaped subgraph based on the shortest path distance. However, DCGNN also simply sums all the node embeddings to get graph representations.

%% file: sec_preliminaries.tex
\section{PRELIMINARIES}
\subsection{Notations and Definitions}\label{sec:nd}
We begin with the notations and definitions used in this paper. We consider an undirected labeled graph $\mathcal{G}=(\mathcal{V},\mathcal{E}, l)$, where $\mathcal{V}=\left\lbrace v_1,\ldots,v_n\right\rbrace $ is a set of vertices (nodes), $\mathcal{E}$ is a set of edges, and $l: \mathcal{V}\rightarrow \Sigma$ is a function that assigns labels from a label alphabet $\Sigma$ to nodes. Without loss of generality, $\lvert\Sigma\rvert\leq |\mathcal{V}|$. An edge $e$ is denoted by two nodes $uv$ that are connected to it. The adjacency matrix of nodes is denoted by $\mathbf{A}\in \mathbb{R}^{n\times n}$ with $a_{i, j}=a_{j, i}, a_{i, i}=0$. The degree matrix $\mathbf{D}$ is a diagonal matrix associated with $\mathbf{A}$ with $d_{i, i}=\sum_j a_{i, j}$. For a graph with self-loops on each node, its adjacency matrix is defined as $\mathbf{\widetilde{A}}=\mathbf{A}+\mathbf{I}$. And the corresponding degree matrix is $\mathbf{\widetilde{D}}$. The indicator function is denoted by $\mathds{1}(x)$. $\mathcal{N}_v$ denotes the one-hop neighbors of node $v$ (excluding itself). $\parallel$~is the concatenation operator. We represent the set of feature vectors of a node's neighbors as a multiset $\ldblbrace \cdot\rdblbrace$ (a set with possibly the same element multiple times).

The homophily ratio $\alpha_v$~\cite{pei2020geom} of a node $v$ in a graph is defined as follows:
\begin{equation}
\label{eqn:v_homo}
\alpha_v=\frac{\sum_{u\in\mathcal{N}_v}\mathds{1}\left(l(u)=l(v)\right)}{\left| \mathcal{N}_v\right|}
\end{equation}

We can see that the larger the $\alpha_v$ value, the more neighboring nodes have the same label as node $v$. For the whole graph, its homophily ratio $\beta$ is computed as follows:
\begin{equation}
\label{eqn:g_homo}
\beta=\frac{1}{\left| \mathcal{V}\right| }\sum_{v\in\mathcal{V}}\alpha_v
\end{equation}
the larger the $\beta$ value, the stronger the homophily of the graph, i.e., the more nodes in the graph have the same label. 



\subsection{Graph Neural Networks}\label{sec:gnn}
Graph Neural Networks (GNNs) use the graph structure and the node features to learn the embeddings or representations of nodes or the entire graph. A majority of GNNs~\cite{gilmer2017neural,defferrard2016convolutional,kipf2016semi} follow the message passing strategy, i.e., iteratively updating the embedding of a node by aggregating those of its neighboring nodes. After $k$ iterations, the embedding of a node captures information from both the graph structure and all its neighbors' embeddings in its $k$-hop neighborhood. In the $k$-th layer, GNNs have two operators:  $\mbox{AGGREGATE}^{(k)}(\cdot)$ and $\mbox{COMBINE}^{(k)}(\cdot)$, whose definitions are given as follows:
\begin{equation}
\label{eqn:agg_com}
\begin{aligned}
\mathbf{h}_{\mathcal{N}_v}^{(k)}&=\mbox{AGGREGATE}^{(k)}\left(\ldblbrace \mathbf{h}_u^{(k-1)}, \forall u\in\mathcal{N}_v\rdblbrace \right)\\
\mathbf{h}_v^{(k)}&=\mbox{COMBINE}^{(k)}\left(\mathbf{h}_v^{(k-1)}, \mathbf{h}_{\mathcal{N}_v}^{(k)}\right)
\end{aligned}
\end{equation}
where $\mathbf{h}_v^{(k)}$ is the embedding of node $v$ at the $k$-th layer. Usually, $\mathbf{h}_v^{(0)}$ is initialized as the node feature vector $\mathbf{x}_v$ (one-hot encoding of the node label or the continuous attributes of node). $\mathbf{h}_{\mathcal{N}_v}^{(k)}$ is the aggregation of the embeddings of neighboring nodes of $v$.



After learning node embeddings in the $k$-th layer, GNNs perform a readout function to aggregate all the node embeddings to generate an embedding for the entire graph. It can be computed as follows:
\begin{equation}
\label{eqn:readout}
\mathbf{h}_\mathcal{G}=\readout\left(\ldblbrace\mathbf{h}_v^{(k)}, \forall v\in\mathcal{G}\rdblbrace \right)
\end{equation}
where $\readout(\cdot)$ should be a permutation- and size-invariant function of node embeddings such as summation or other advanced techniques, e.g.,  DeepSets~\cite{zaheer2017deep}.

%% file: sec_methods.tex
\section{IHGNN}

\begin{figure*}[!htp]
 \centering	
 \includegraphics[width=0.85\textwidth]{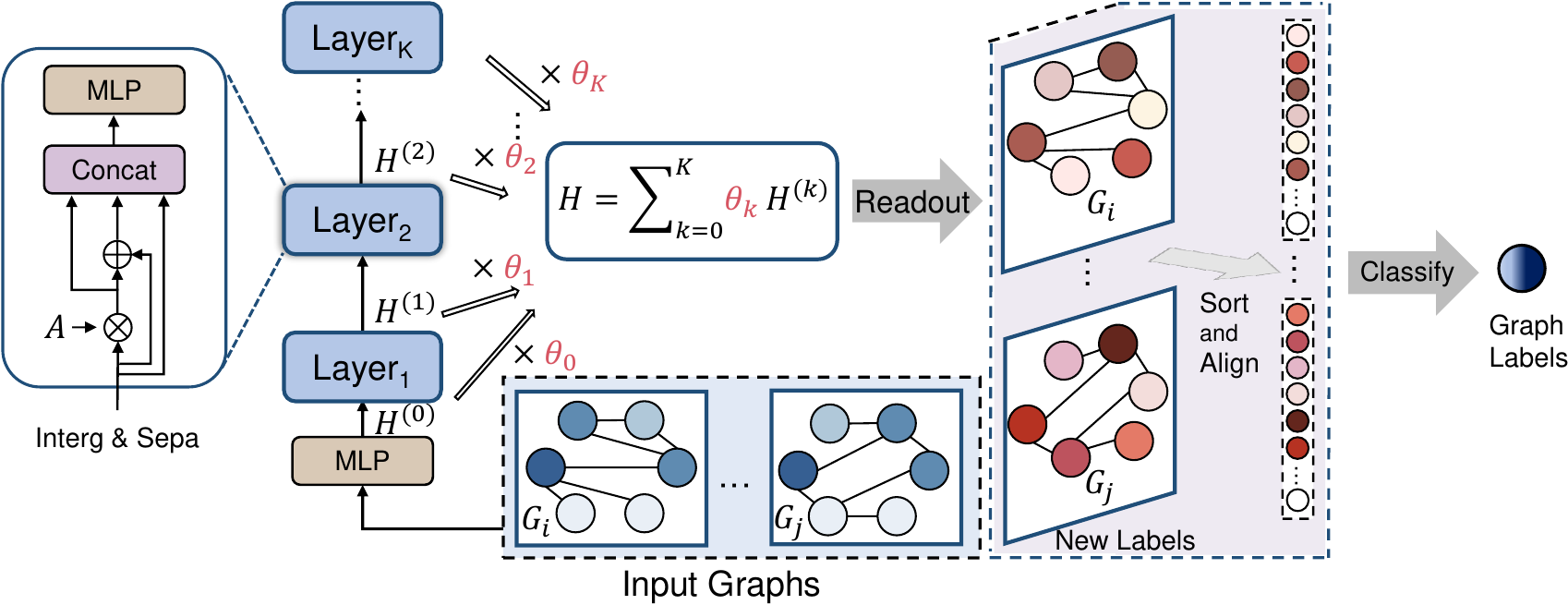}
 \caption{The pipeline of our model IHGNN.
 Each IHGNN layer consists of the integration and separation of ego- and neighbor-embeddings. All layers are adaptively aggregated to form the node embeddings $\mathbf{H}$. 
The process of generating node embeddings $\mathbf{H}$ is analogous to applying the 1-WL algorithm for $K$ iterations. The graph-level readout function utilizes the last column of $\mathbf{H}$ as the final continuous labels of the nodes. Subsequently, We sort and align the nodes based on the final continuous labels.
 Dummy nodes are appended to ensure equal graph size.
	}
	\label{flow_chart}
\end{figure*}

One of the major motivations of our proposed method is to consider the fact that the heterophily property exists in real-world graphs. In molecules, 
 atoms tend to form chemical bonds with both the same and different kinds of atoms. Thus, nodes in such molecule graphs have both homophily and heterophily. 
For a graph with varying node homophily ratios, a GNN that is designed only for one mode cannot model the intricate relations between nodes. Furthermore, emphasizing a certain graph homophily ratio rather than adapting to numerous graphs with varying homophily ratios underperforms in graph classification. So in this paper, we design a novel GNN that can simultaneously consider both modes at node-level and graph-level.
The overall framework of IHGNN is shown in Fig.~\ref{flow_chart}. 
Next, we describe our designs in more detail.

\subsection{Integration and Separation of Ego- and Neighbor-embeddings}
GIN~\cite{xu2018powerful} argues that the sum aggregator has more expressive power than the mean and max aggregators over a multiset. Here, we also use the sum aggregator as our aggregation function, $\mbox{AGGREGATE}^{(k)}(\cdot)$:
\begin{equation}
\label{eqn:neig}
\mathbf{h}_{\mathcal{N}_v}^{(k)}=\sum_{u\in\mathcal{N}_v}\mathbf{h}_u^{(k-1)}
\end{equation}

For graphs with high node homophily ratios, integrating the ego- and neighbor-embeddings of nodes such as GCN~\cite{kipf2016semi} has been proven to be effective in node classification. For graphs with low node homophily ratios, separating the ego- and neighbor-embeddings of nodes such as H$_2$GCN~\cite{zhu2020beyond} has been proven to be effective in node classification. We infer that for graphs with varying node homophily ratios, using both the integration and separation of the ego- and neighbor-embeddings of nodes might be effective even in graph classification. To consider both the integration and separation of the ego- and neighbor-embeddings of nodes, we develop $\mbox{COMBINE}^{(k)}(\cdot)$ as follows:
\begin{equation}
\label{eqn:comb1}
\mathbf{h}_v^{(k)}=\mbox{MLP}^{(k)}\left(\mathbf{h}_v^{(k-1)}\parallel\mathbf{h}_{\mathcal{N}_v}^{(k)}\parallel\left(\mathbf{h}_v^{(k-1)}+\mathbf{h}_{\mathcal{N}_v}^{(k)}\right)\right)
\end{equation}
where $\mbox{MLP}^{(k)}$ denotes the Multi-layer Perceptrons for the $k$-th layer. We concatenate each node's ego-embedding,  neighbor-embedding, and their summation, and use MLP to generate a new ego-embedding for each node.

As indicated by Zhu et al.~\cite{zhu2020beyond}, in graphs with a low homophily ratio, the node label distribution contains more information at the higher orders of the adjacency matrix than the lower orders of the adjacency matrix. Therefore, the aggregation of the intermediate node embeddings from different layers increases the performance of GNNs. They concatenate all the intermediate node embeddings from different layers. However, the concatenation is not adaptive to different datasets. In this paper, we propose to adaptively aggregate all these intermediate node embeddings by a Single-Layer Perceptron (SLP) as follows: 
 \begin{equation}
  \label{eqn:slp}
 \mathbf{h}_v = \sum_{k=0}^{K}\theta_k\mathbf{h}_v^{(k)}
 \end{equation}
where $K$ is the number of layers and $\theta_k$ is the learning parameter of SLP for the $k$-th layer.

\subsection{Graph-level Readout Function}


 
In the previous section, we obtained the representation or embedding for each node (equation~(\ref{eqn:slp})). Many variants of GNNs use the summation as the final graph-level readout function. However, as for graphs with varying node homophily ratios, the summation of all the node embeddings ignores the characteristic (homophily ratio) of each node, which makes the following MLP classifier not aware of different node homophily ratios. This will decrease the performance of GNNs. To mitigate this problem, we propose to concatenate all the nodes’ aggregated embeddings as shown in equation~(\ref{eqn:slp}), i.e., making GNNs aware of nodes with different homophily ratios. However, the concatenation function does not satisfy the conditions of the graph-level readout function, i.e., permutation-invariance and size-invariance.
\begin{figure*}[t]
\centering
\subfigure[$\mathcal{G}_1$]
{
 	\begin{minipage}[b]{.17\linewidth}
        \centering
        \includegraphics[scale=0.53]{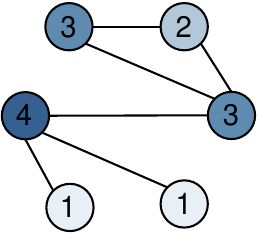}
    \end{minipage}
}
\subfigure[$\mathcal{G}_1$: neighborhood aggregation]
{
 	\begin{minipage}[b]{.27\linewidth}
        \centering
        \includegraphics[scale=0.53]{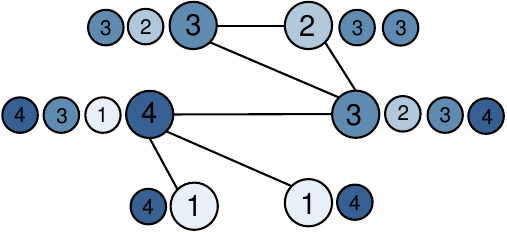}
    \end{minipage}
}
\subfigure[$\mathcal{G}_1$: label hashing across graphs]
{
 	\begin{minipage}[b]{.22\linewidth}
        \centering
        \includegraphics[scale=0.43]{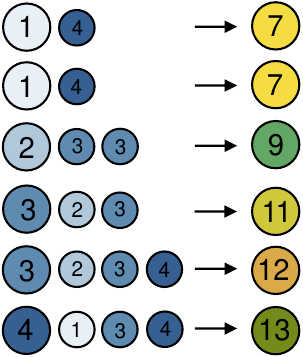}
    \end{minipage}
}
\subfigure[$\mathcal{G}_1$: new labels]
{
 	\begin{minipage}[b]{.17\linewidth}
        \centering
        \includegraphics[scale=0.53]{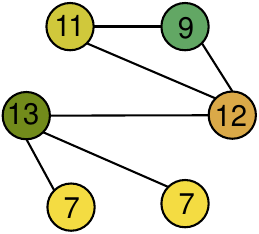}
    \end{minipage}
}
\subfigure[$\mathcal{G}_2$]
{
 	\begin{minipage}[b]{.17\linewidth}
        \centering
        \includegraphics[scale=0.53]{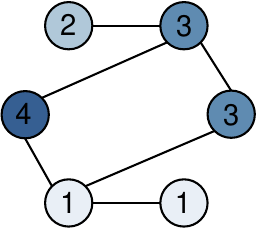}
    \end{minipage}
}
\subfigure[$\mathcal{G}_2$: neighborhood aggregation]
{
 	\begin{minipage}[b]{.27\linewidth}
        \centering
        \includegraphics[scale=0.53]{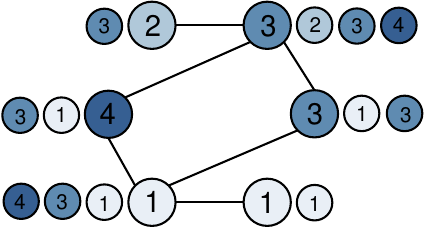}
    \end{minipage}
}
\subfigure[$\mathcal{G}_2$: label hashing across graphs]
{
 	\begin{minipage}[b]{.22\linewidth}
        \centering
        \includegraphics[scale=0.43]{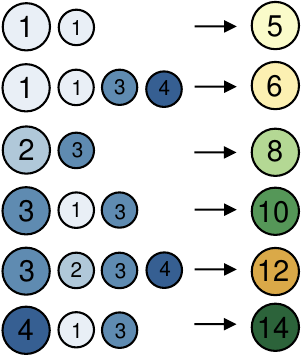}
    \end{minipage}
}
\subfigure[$\mathcal{G}_2$: new labels]
{
 	\begin{minipage}[b]{.17\linewidth}
        \centering
        \includegraphics[scale=0.53]{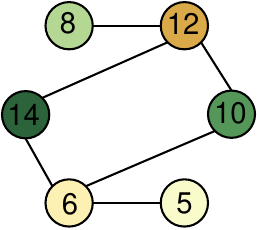}
    \end{minipage}
}

	\caption{Illustration of one iteration of the first order Weisfeiler-Lehman isomorphism test (1-WL) for graphs. (b) and (f) show that for each node in $\mathcal{G}_1$ and $\mathcal{G}_2$, 1-WL aggregates/concatenates its label and the labels of its neighboring nodes as a string. (c) and (g) show that 1-WL uses an injective hash function to project the label string to a new integer label. For (a) and (e), $\Sigma=\{1,2,3,4\}$. For (d) and (h), $\Sigma=\{5,6,7,8,9,10,11,12,13,14\}$.}
	\label{fig:wlcoloring}
\end{figure*}

\textbf{Satisfying permutation-invariance:} To solve the problem of permutation-invariance, we propose to align nodes such that nodes from different graphs are sorted and aligned in the same order, which is consistent across different graphs. Now the ordering in which we sort the nodes so that they are aligned across different graphs becomes important. 
Note that the 1-WL~\cite{leman1968reduction} algorithm defines an ordering based on the node structural identity and this ordering is consistent across different graphs. For example, in Fig.~\ref{fig:wlcoloring}(c) and (g), the 1-WL after one iteration will sort the nodes in $\mathcal{G}_1$ in an order of $7,7,9,11,12,13$ and sort the nodes in $\mathcal{G}_2$ in an order of $5,6,8,10,12,14$. The two nodes with label $12$ in $\mathcal{G}_1$ and $\mathcal{G}_2$ have the same position in the two orders because they have the same structural identity as indicated in Fig.~\ref{fig:wlcoloring}(a) (the right-most node with label 3) and Fig.~\ref{fig:wlcoloring}(e) (the upper-right node with label 3), respectively. It has been indicated in GCN~\cite{kipf2016semi} that the $\mbox{AGGREGATE}^{(k)}(\cdot)$ and $\mbox{COMBINE}^{(k)}(\cdot)$ operators are the continuous versions of the aggregate and hash operators in the 1-WL. Thus, nodes in each graph can be sorted by their final embeddings, which can be considered as the continuous 1-WL labels.
Please note the distinctions in colors and arrangement of the new labels between Fig.~\ref{flow_chart} and Fig.~\ref{fig:wlcoloring}.
The discrepancy arises from the fact that the generation process of node embeddings $\mathbf{H}$ in Fig.~\ref{flow_chart} can be seen as applying the 1-WL algorithm for $K$ iterations while in Fig.~\ref{fig:wlcoloring} the 1-WL algorithm runs for one iteration.

\textbf{Satisfying size-invariance:} To solve the problem of size-invariance, we use the size (e.g., $m$) of the graph that has the largest number of nodes as the common size. For graphs whose sizes are less than $m$, we concatenate them with dummy nodes to make their sizes equal to $m$. The features of the dummy nodes are set to zero vectors so that they do not contribute to learning the associated parameters. After aligning the nodes from different graphs and making different graphs the same size, the graph-level readout function is given as follows:
 \begin{equation}
 \label{eqn:grd}
\mathbf{h}_\mathcal{G} = \mathbf{h}_{\sigma_{v_1}}\parallel\mathbf{h}_{\sigma_{v_2}}\parallel\cdots\parallel\mathbf{h}_{\sigma_{v_{m-w}}}\parallel\underbrace{\mathbf{0}\parallel\cdots\parallel\mathbf{0}}_w
\end{equation}
where $m-w$ denotes the number of nodes in a graph that does not have the largest node number and $\sigma_{v_1}, \sigma_{v_2},\ldots,\sigma_{v_{m-w}}$ is an order on nodes defined by the continuous 1-WL labels after $K$ iterations. Finally, we input $\mathbf{h}_\mathcal{G}$ into MLP followed by a softmax layer for graph classification.

\begin{algorithm2e}
	\KwIn{A set of  graphs $\{\mathcal{G}_1, \mathcal{G}_2, \ldots, \mathcal{G}_n\}$ and their labels $\mathcal{Y}=\{y_1,y_2,\ldots,y_n\}$, the number $K$ of layers, batch size $b$, the number of epochs $E$}
	\KwOut{Classification accuracy}
	Initialize the parameters of all the neural networks\;
	\While{\upshape epoch $\leq E$}{
		\ForEach{\upshape batch of graphs $\{\mathcal{G}_{\gamma_1}, \mathcal{G}_{\gamma_2}, \ldots, \mathcal{G}_{\gamma_b}\}$ in the training data}{
			\tcc{$\gamma_1, \gamma_2, \ldots, \gamma_b$ denote the indexes in this batch of graphs}
			\ForEach{\upshape graph $\mathcal{G}$ in the batch }{
				$\mathbf{H}^{(0)}\leftarrow \mbox{MLP}^{(0)}\left(\mathbf{X}\right)$\tcc*[r]{Each row of $\mathbf{X}$ represents the one-hot encoding of each node label.}
				\For{$k\leftarrow 1$ \KwTo $K$}{$\mathbf{H}^{(k)}\leftarrow\mbox{MLP}^{(k)}(\mathbf{H}^{(k-1)}\parallel$
                $\mathbf{A}\cdot\mathbf{H}^{(k-1)}\parallel\left(\mathbf{H}^{(k-1)}+\mathbf{A}\cdot\mathbf{H}^{(k-1)}\right))$\tcc*[r]
                    {$\mathbf{A}$ is the adjacency matrix of the graph $\mathcal{G}$.}
				}
				$\mathbf{H}\leftarrow\sum_{k=0}^K\theta_k\mathbf{H}^{(k)}$\;
				Sort nodes in ascending order w.r.t. the continuous 1-WL labels\;
				Generate the graph embedding $\mathbf{h}_\mathcal{G}$ using equation~(\ref{eqn:grd}) and input it into MLP for graph classification\;
			}
			Back-propagate to update the parameters of all the neural networks\;
		}
	}
	\Return{\upshape{Classification accuracy}}\;
	\caption{IHGNN}
	\label{alg:IHGNN}
\end{algorithm2e}

\begin{table*}[!htb]

\centering
\caption{Comparison of classification accuracy (mean $\pm$ standard deviation) of IHGNN to other GNNs on the benchmark datasets. In general, IHGNN outperforms the others on 7 out of 9 datasets.
 The best results per benchmark are highlighted in \textbf{bold} and runner-ups are highlighted in gray. N/A means the node labels are not available in social datasets and the node degrees are used as labels, and OOM means Out of GPU Memory.
 $^\dag$ means the results are adopted from \cite{errica2019fair}.
 }
	\label{tab:classification2}

\begin{tabular}{lccccccccc}
\toprule
                         & \textbf{KKI}       & \textbf{DD}        & \textbf{COLLAB}    & \textbf{IMDB-B}    & \textbf{DHFR\_MD}  & \textbf{BZR\_MD}   & \textbf{MUTAG}     & \textbf{PROTEINS}  & \textbf{IMDB-M}    \\
\textbf{Hom. ratio} $\beta$     & 0.00$\pm$0.00 & 0.07$\pm$0.02 & 0.30$\pm$0.33 & 0.46$\pm$0.27 & 0.51$\pm$0.07 & 0.60$\pm$0.08 & 0.65$\pm$0.14 & 0.66$\pm$0.18 & 0.72$\pm$0.32 \\
\textbf{\#Graph numbers}     & 83 & 1178 & 5000 & 1000 & 393 & 306 & 188 & 1113 & 1500 \\
\textbf{\#Average nodes}  & 26.96              & 284.32             & 74.49              & 19.77              & 23.87              & 21.30              & 17.93              & 39.06              & 13.00              \\
\textbf{\#Average edges} & 48.42              & 715.66             & 2457.78            & 96.53              & 283.01             & 225.06             & 19.79              & 72.82              & 65.93              \\
\textbf{\#Node labels}    & 190                & 82                 & N/A                & N/A                & 7                  & 8                  & 7                  & 3                  & N/A                \\
\midrule
\midrule
PPGNs                    & 57.9$\pm$1.6           & OOM                & OOM                & 71.7$\pm$3.9           & 67.2$\pm$3.4           & 70.9$\pm$10.8          & 86.7$\pm$6.7           & OOM                & 49.3$\pm$3.1          \\
$k$-GNNs                   & 58.8$\pm$1.4           & 78.4$\pm$6.3           & OOM                & 71.0$\pm$4.1           & 67.4$\pm$5.0           & 71.8$\pm$1.0           & 82.9$\pm$6.8           & 76.0$\pm$3.0           & 49.3$\pm$3.1          \\
GNNML                    & 53.3$\pm$12.2          & 77.0$\pm$3.1$^\dag$           & 78.5$\pm$1.7           & 69.6$\pm$3.4           & 67.9$\pm$1.1           & 69.6$\pm$9.6           & 86.2$\pm$7.8           & 74.4$\pm$2.2           & 43.5$\pm$2.8           \\
\midrule
Graphormer               & 51.8$\pm$10.7          & \cellcolor{gray!25}82.9$\pm$7.3                  & 72.5$\pm$2.1           & 71.5$\pm$7.3           & 66.4$\pm$4.0           & 58.8$\pm$8.1           & 86.8$\pm$6.3           & 73.1$\pm$3.9           & 48.6$\pm$3.2           \\
TokenGT                  & 55.4$\pm$5.2           & \textbf{83.0$\pm$8.2}                  & 66.1$\pm$3.0           & 66.7$\pm$3.5           & 67.9$\pm$1.1           & 60.4$\pm$6.9           & 84.0$\pm$6.0           & 71.2$\pm$3.3           & 46.9$\pm$1.7           \\
\midrule
DiffPool                 & 49.3$\pm$11.2          & 75.0$\pm$3.5$^\dag$           & 68.9$\pm$2.0$^\dag$           & 68.4$\pm$3.3$^\dag$           & 64.7$\pm$4.7           & 67.3$\pm$3.6           & 80.6$\pm$5.4           & 73.7$\pm$3.5$^\dag$           & 45.8$\pm$3.7           \\
DGCNN                    & 50.4$\pm$5.6           & 76.6$\pm$4.3$^\dag$           & 71.2$\pm$1.9$^\dag$           & 69.2$\pm$3.0$^\dag$           & 65.0$\pm$4.4           & 65.1$\pm$4.6           & 82.4$\pm$7.6           & 72.9$\pm$3.5$^\dag$           & 45.6$\pm$3.5           \\
GIN                      & 60.1$\pm$15.0           & 75.3$\pm$2.9$^\dag$           & 75.6$\pm$2.3$^\dag$           & 71.2$\pm$3.9$^\dag$           & 65.9$\pm$5.5           & 62.2$\pm$7.6           & 83.4$\pm$6.4           & 73.3$\pm$4.0$^\dag$           & 48.5$\pm$3.3$^\dag$           \\
H$_2$GCN                      & 60.3$\pm$15.1                   & 79.0$\pm$4.0                   & \textbf{79.5$\pm$1.7}                   &  73.0$\pm$4.3                  &  68.7$\pm$2.0                  &   71.5$\pm$10.8                 &  82.4$\pm$5.9                  &  75.6$\pm$2.6                  &  \cellcolor{gray!25}51.1$\pm$2.6  \\
GPR-GNN                  &  60.4$\pm$11.3                  &  58.8$\pm$5.0                 &  52.0$\pm$5.9                &  58.0$\pm$7.0                  &   67.9$\pm$1.1                 &  55.3$\pm$7.3                  &  79.8$\pm$10.9                  &     61.7$\pm$3.4               &  35.9$\pm$3.2                 \\
ACM                      & 59.9$\pm$17.5                   &         77.5$\pm$3.8           &  77.8$\pm$1.5                  &  \cellcolor{gray!25}74.0$\pm$3.3                  &  \cellcolor{gray!25}69.9$\pm$5.1                  &  \cellcolor{gray!25}71.9$\pm$9.3                 &  80.8$\pm$8.1                  & \cellcolor{gray!25}76.2$\pm$3.1                   & 50.7$\pm$2.6                  \\
PGCN                     & 55.0$\pm$15.5          & 77.5$\pm$3.5           & 73.3$\pm$1.6           & 73.5$\pm$6.3           & 67.4$\pm$7.8           & 71.8$\pm$8.8           & 88.4$\pm$2.1           & 75.7$\pm$1.6           & 48.9$\pm$3.0           \\
DCGNN                     & 56.9$\pm$12.2          &  59.3$\pm$1.7          & 70.3$\pm$2.7           &  71.3$\pm$3.7          &    67.9$\pm$1.1        &  56.5$\pm$9.0          & 83.5$\pm$6.4           &  61.7$\pm$3.2          &  49.6$\pm$3.1          \\
PCNet                     &  \cellcolor{gray!25}63.6$\pm$12.9          & 58.7$\pm$3.0           & 72.2$\pm$2.2           &  72.1$\pm$4.7         &   68.5$\pm$1.2         &  52.2$\pm$3.2         & 87.8$\pm$7.2           &   62.0$\pm$5.6         &  47.3$\pm$2.6          \\
\midrule
IHGNN                    & \textbf{67.8$\pm$18.7} & 79.2$\pm$2.5  & \cellcolor{gray!25}79.0$\pm$2.0  & \textbf{74.1$\pm$5.4}  & \textbf{74.3$\pm$6.0}  & \textbf{72.2$\pm$7.8}  & \textbf{89.9$\pm$6.0}  & \textbf{76.2$\pm$2.2}  & \textbf{51.9$\pm$2.5}\\
\bottomrule
\end{tabular}

\end{table*}

\begin{table*}[!htb]

\centering
\caption{Comparison of classification accuracy (mean $\pm$ standard deviation) of IHGNN to its four variants on the benchmark datasets. 
In general, IHGNN outperforms its variants on 8 out of 9 datasets.
The best results per benchmark are highlighted in \textbf{bold}.
The results as \underline{underlined} highlight the lowest variants in the ablation study.
 }
	\label{tab:ablationStudy}

\begin{tabular}{lccccccccc}
\toprule
                         & \textbf{KKI}       & \textbf{DD}        & \textbf{COLLAB}    & \textbf{IMDB-B}    & \textbf{DHFR\_MD}  & \textbf{BZR\_MD}   & \textbf{MUTAG}     & \textbf{PROTEINS}  & \textbf{IMDB-M}    \\

\midrule

IHGNN-w/o-Integ          & 57.3$\pm$16.0          & 78.2$\pm$4.1           & 78.6$\pm$1.7           & \underline{73.3$\pm$4.1}           & 69.7$\pm$5.1           & 67.3$\pm$5.1           & 88.3$\pm$5.2           & \underline{74.6$\pm$3.0}           & \underline{49.5$\pm$2.9}           \\
IHGNN-w/o-Sepa           & 59.0$\pm$18.0          & 78.7$\pm$2.5           & \underline{78.0$\pm$1.4}           & 73.4$\pm$4.4           & 73.5$\pm$4.5           & 71.5$\pm$10.2          & 87.7$\pm$6.0           & 75.3$\pm$2.7           & 50.7$\pm$2.8           \\
IHGNN-w/o-Adapt          & \underline{55.0$\pm$18.9}          & 77.8$\pm$3.4           & 78.2$\pm$1.9           & 73.7$\pm$4.6           & 72.3$\pm$5.2           & 68.0$\pm$9.9           & 89.4$\pm$4.1           & 74.8$\pm$2.8           & 49.9$\pm$3.6           \\
IHGNN-w/o-Concat         & 61.3$\pm$17.0          & \underline{75.5$\pm$3.4}           & \textbf{79.2$\pm$1.4}           & 73.9$\pm$3.2           & \underline{68.7$\pm$8.3}           & \underline{66.9$\pm$10.5}          & \underline{86.8$\pm$6.7}           & 75.5$\pm$2.1           &51.2$\pm$2.1           \\
IHGNN                    & \textbf{67.8$\pm$18.7} & \textbf{79.2$\pm$2.5}  & 79.0$\pm$2.0  & \textbf{74.1$\pm$5.4}  & \textbf{74.3$\pm$6.0}  & \textbf{72.2$\pm$7.8}  & \textbf{89.9$\pm$6.0}  & \textbf{76.2$\pm$2.2}  & \textbf{51.9$\pm$2.5}\\
\bottomrule
\end{tabular}

\end{table*}

\subsection{Computational Complexity Analysis}

The pseudo-code of IHGNN is given in Algorithm 1. Line~1 initializes the parameters of all the neural networks. At line 5, for each graph $\mathcal{G}$ in a batch of training data, we use MLP to transform the one-hot encoding $\mathbf{X}\in\mathbb{R}^{n\times c}$ of node labels to the embeddings
$\mathbf{H}^{(0)}\in\mathbb{R}^{n\times r}$, where $c$ is the number of classes, $r$ is the dimension of the embedding. Assume that every graph has the same number $n$ of nodes and the number of neurons of the hidden and output layers of MLP is equal. The time complexity of line~5 is $\mathcal{O}(n c r + n r r )$. Lines 6--7 update each node embedding iteratively by the $\mbox{AGGREGATE}^{(k)}(\cdot)$ and $\mbox{COMBINE}^{(k)}(\cdot)$ operators in equation~(\ref{eqn:neig}) and equation~(\ref{eqn:comb1}). The time complexity is $\mathcal{O}\left( K\left( \mbox{vol}(\mathcal{G}) r +n r +n r^2\right) \right)$, where $\mbox{vol}(\mathcal{G})=\sum_id_{i,i}$ is the volume of the graph $\mathcal{G}$. In line 8, we use SLP to adaptively aggregate all the intermediate node embeddings, of which the time complexity is $\mathcal{O}(nrK)$. In line 9, we sort nodes in ascending order according to their continuous 1-WL labels.
The time complexity is $\mathcal{O}(n \log(n))$. At line 10, we generate the graph embedding $\mathbf{h}_\mathcal{G}$ and input it into MLP for graph classification. The time complexity is $\mathcal{O}(n r r' + n r' c)$, where $r'$ is the number of neurons of the hidden layer of MLP for graph classification. Finally, line 11 utilizes the back-propagation strategy to update the parameters of all the neural networks. Since $r'$ is in the same scale as $r$, the total time complexity of IHGNN in one epoch on a batch $b$ of graphs is bounded by 
$\mathcal{O}\left( b\left( n( K r^2+(c+K) r)+\mbox{vol}(\mathcal{G}) K r+ n\log(n)\right) \right)$.

%% file: sec_experiments.tex
\section{Experimental Evaluation}\label{sec:experiments}

\subsection{Experimental Setup}

Since IHGNN is a method specifically designed for graph classification tasks, we choose graph classification methods such as PPGNs~\cite{maron2019provably}, $k$-GNNs~\cite{morris2019weisfeiler}, GNNML~\cite{balcilar2021breaking}, DiffPool~\cite{ying2018hierarchical}, DGCNN~\cite{zhang2018end}, GIN~\cite{xu2018powerful}, PGCN~\cite{pasa2022polynomial}, and DCGNN~\cite{wu2024dcgnn} as baselines. We also compare with some node classification methods considering the heterophily such as H$_2$GCN~\cite{zhu2020beyond}, GPR-GNN~\cite{chien2020adaptive},  ACM~\cite{luan2022revisiting}, and PCNet~\cite{li2024pc}. A sum pooling function and a 2-layer MLP are added to convert them into graph classification methods. Furthermore, we compare with two transformer-based models, i.e., Graphormer~\cite{ying2021transformers} and TokenGT~\cite{kim2022pure}.

We adopt 9 benchmark graph datasets to evaluate each method. 
For a fair comparison, we use the cross-validation procedure proposed in~\cite{errica2019fair} for each method. The procedure uses a 10-fold cross-validation for model evaluation and an inner holdout method with 90\%/10\% training/validation split for model selection. 
We use the same data splits as used in~\cite{errica2019fair}. If a specific dataset is not used in~\cite{errica2019fair}, we use the original code of~\cite{errica2019fair} to generate a data split for this specific dataset. The mean accuracy and standard deviation of all models are reported on the test sets over 10 folds.
For all the comparison methods, we set their hyperparameters according to their original papers. For IHGNN, we train for 350 epochs using Adam~\cite{kingma2014adam} optimization method with a learning rate selected from $\left\lbrace 0.01, 0.001, 0.0001\right\rbrace$. The batch size is selected from $\left\lbrace 32, 64, 128\right\rbrace $. The number of layers is selected from $\left\lbrace 3, 4, 5\right\rbrace $. The dropout ratio is set to 0.5, the number of neurons of the hidden and output layers of all the MLPs in the $\mbox{COMBINE}^{(k)}(\cdot)$ operator is set to 32, and the number of neurons of the hidden layer of the final MLP for graph classification is set to 128. We run all the experiments on a server with a dual-core Intel(R) Xeon(R) Gold 6226R CPU @ 2.90GHz, 256 GB memory, an Nvidia GeForce RTX 3090 GPU, and an Ubuntu 18.04.1 LTS operating system. We make our code publicly available at \url{https://anonymous.4open.science/r/IHGNN-F070}.

\subsection{Results}

\subsubsection{Classification Accuracy}

The classification accuracy of each method on the 9 benchmark datasets is shown in Table~\ref{tab:classification2}. 
We observe that IHGNN achieves the best results on 7 out of 9 datasets while securing the second-best performance on the COLLAB dataset.
Specifically, on the KKI dataset, IHGNN has a gain of 6.6\% over the runner-up method PCNet and a gain of 37.5\% over the worst method DiffPool. 
We can see that the graph homophily ratio of the KKI dataset is zero, which means the KKI dataset is heterophilous.
All the comparison methods that do not consider the separation of the ego- and neighbor-embeddings of nodes lead to inferior performance. 
On datasets DHFR\_MD, MUTAG, and IMDB-M that tend to be more homophilous, IHGNN also has a gain of 6.3\%, 1.7\%, and 1.6\% over the runner-up method, respectively.
The superior performance of IHGNN on graphs with both homophily and heterophily is attributed to our thoughtful integration of heterophily in both node and graph representation learning processes. 

Regarding the comparison methods, PPGNs, $k$-GNNs, and GNNML utilize high-order graph structures and are supposed to be more expressive than the 1-WL methods. However, we can see that they only achieve fairly good results on classifying graphs with varying homophily ratios and encounter out-of-memory problems on large graphs due to their high computational costs.
Graphormer and TokenGT exhibit subpar performance across all datasets except the DD dataset. We believe that might be due to the lack of pertaining and the overfitting problem of transformers on small datasets. Notably, IHGNN demonstrates the best performance among all the GNN-based methods on the DD dataset.
In addition, as expected, the top-performing and runner-up methods all take the heterophily into account, except on the DD dataset, where Transformer-based method TokenGT is the best and Graphormer is the runner-up.


\subsubsection{Ablation Study}
In this section, we perform an ablation study on each strategy used in IHGNN: a) IHGNN without the integration of the ego- and neighbor-embeddings of nodes (IHGNN-w/o-Integ); b) IHGNN without the separation of the ego- and neighbor-embeddings of nodes (IHGNN-w/o-Sepa); c) IHGNN without using the SLP for adaptive aggregation of node embeddings from each layer (using concatenation instead, IHGNN-w/o-Adapt); and d) IHGNN without using the concatenation of all the node embeddings as the final graph-level readout function (using summation instead, IHGNN-w/o-Concat). The classification accuracy is reported in Table~\ref{tab:ablationStudy}. We observe that IHGNN outperforms IHGNN-w/o-Integ, IHGNN-w/o-Sepa, and IHGNN-w/o-Adapt on all the datasets. Compared with IHGNN-w/o-Concat, IHGNN also achieves the best results on 8 out of 9 datasets. 

Each strategy has a different contribution to the overall performance of IHGNN on each dataset. Among the four variants, IHGNN-w/o-Concat and IHGNN-w/o-Integ attain the lowest accuracy the most frequently, indicating the readout function and the integration of the ego- and neighbor-embedding contribute the most. This also validates that
our design to 
adapt to different graph homophily ratios in the datasets is
pivotal to our graph classification tasks.


\subsubsection{Parameter Sensitivity}
\begin{figure}[!htb]
\centering
\includegraphics[scale=0.25]{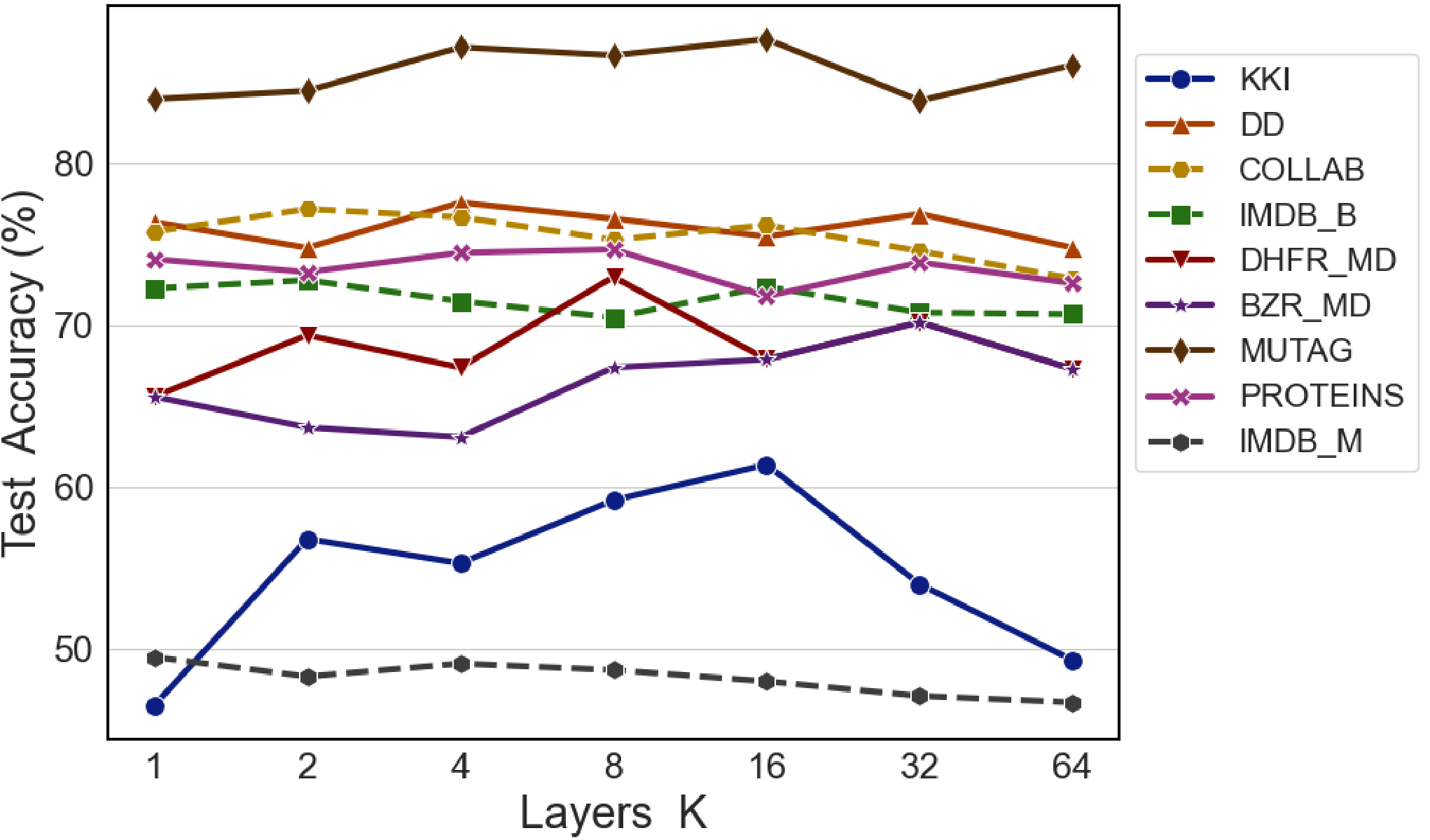}
\label{figure}
\caption{Parameter sensitivity of IHGNN over 10 different random seeds on 9 benchmark datasets. }
\label{ParameterSensitivity}
\end{figure}

We investigate how the number of layers affects the performance of IHGNN on the 9 benchmark datasets.
The value of the layer $K$ varies in the range $[1, 2, 4, 8, 16, 32, 64]$. 
Each experiment is run 10 times, with different random seeds. The average classification accuracy on the test sets over the 10 times is shown in Fig.~\ref{ParameterSensitivity}.
We observe that IHGNN demonstrates outstanding performance even with a large number of layers, which we conjecture is owing to the preservation of the inter-layer information by the adaptive aggregation design. In general, across 6 out of 9 datasets, IHGNN gets optimal results when the layer $K$ is greater or equal to 4. Specifically, IHGNN performs the best on the KKI and MUTAG datasets when $K=16$, and on the BZR\_MD dataset when $K=32$.


%% file: sec_conclusion.tex
\section{Conclusion}\label{conclusion}

In this paper, we have developed a new GNN model called IHGNN for graphs with varying node homophily ratios. Many existing GNNs fail to generalize to this setting and thus underperform on such graphs. Our designs distinguish not only between the integration and separation of the ego- and neighbor-embeddings of nodes in the $\mbox{COMBINE}(\cdot)$ operator used in GNNs but also between each node embedding in the graph-level readout function. In addition, we adaptively aggregate node embeddings from different layers. These designs empowers IHGNN's performance. Experiments show that IHGNN outperforms other state-of-the-art GNNs on graph classification. In the future, we would like to apply these designs on GNNs that perform message passing between higher-order relations in graphs.